\documentclass[runningheads]{llncs}
\usepackage{graphicx}
\usepackage{comment}
\usepackage{amsmath,amssymb} 
\usepackage{color}

\usepackage{epsfig}
\usepackage{graphicx}
\usepackage{amsfonts,bm}
\usepackage{graphics}
\usepackage{float}
\usepackage{array}
\usepackage{booktabs}
\usepackage{multirow}
\usepackage[table,xcdraw]{xcolor}
\usepackage{lipsum}
\usepackage{hhline}
\usepackage{makecell}

\usepackage{caption}
\usepackage{subcaption}
\usepackage{setspace}
\usepackage{multicol}

\newcommand{\muv}{{\bm{\mu}}}

\newcommand{\zv}{\mathbf{z}}
\newcommand{\xv}{\mathbf{x}}

\newcommand{\fv}{\mathbf{f}}

\newcolumntype{?}{!{\vrule width 2pt}}

\newcommand{\Sigmam}{\bm{\Sigma}}

\newcommand{\Xm}{\mathbf{X}}
\newcommand{\Zm}{\mathbf{Z}}

\newcommand{\PreserveBackslash}[1]{\let\temp=\\#1\let\\=\temp}
\newcolumntype{C}[1]{>{\PreserveBackslash\centering}p{#1}}
\newcolumntype{R}[1]{>{\PreserveBackslash\raggedleft}p{#1}}
\newcolumntype{L}[1]{>{\PreserveBackslash\raggedright}p{#1}}

\begin{document}
\pagestyle{headings}
\mainmatter
\def\ECCVSubNumber{4912}  

\title{BioMetricNet: deep unconstrained face verification through learning of metrics regularized onto Gaussian distributions} 

\titlerunning{BioMetricNet}
%
\author{Arslan Ali\orcidID{0000-0003-0282-0726} \and
Matteo Testa\orcidID{0000-0003-2628-6433} \and \\
Tiziano Bianchi\orcidID{0000-0002-3965-3522}\and
Enrico Magli\orcidID{0000-0002-0901-0251}}
\authorrunning{Ali. A et al.}
%

\institute{ Department of Electronics and Telecommunications \\  Politecnico di Torino, Italy \\
\email\{arslan.ali, matteo.testa, tiziano.bianchi, enrico.magli\}@polito.it}
\maketitle

\begin{abstract}
We present BioMetricNet: a novel framework for deep unconstrained face verification which learns a regularized metric to compare facial features. 
Differently from popular methods such as FaceNet, the proposed approach does not impose any specific metric on facial features; instead, it shapes the decision space
by learning a latent representation in which matching and non-matching pairs are mapped onto clearly separated and well-behaved target distributions. In particular, the network  jointly learns the best feature representation, and the best metric that follows the target distributions, to be used to discriminate face images.
In this paper we present this general framework, first of its kind for facial verification, and tailor it to Gaussian distributions. This choice enables the use of a  simple linear decision boundary that can be tuned to achieve the desired trade-off between false alarm and genuine acceptance rate, and leads to a loss function that can be written in closed form.
Extensive analysis and experimentation on publicly available datasets such as Labeled Faces in the wild (LFW), Youtube faces (YTF), Celebrities in Frontal-Profile in the Wild (CFP), and challenging datasets like cross-age LFW (CALFW), cross-pose LFW (CPLFW), In-the-wild Age Dataset (AgeDB) show a significant performance improvement and confirms the effectiveness and superiority of BioMetricNet over existing state-of-the-art methods.
\end{abstract}

\section{Introduction}
\label{sec:introduction}
Over the last few years, huge progress has been made in the deep learning community. Advances in convolutional neural networks (CNN) have led to unprecedented accuracy in many computer vision tasks. 
One of those that have attracted computer vision researchers since its inception is being able to recognize a person from a picture of their face. This task, which has countless applications, is still far to be marked as a solved problem. Given a pair of (properly aligned)  face images, the goal is to make a decision on whether they represent the same person or not. 

Early attempts in the field required the design of handcrafted features that could capture the most significant traits that are unique to each person. Furthermore, they had to be computed from a precisely aligned and illumination normalized picture. The complexity of handling the non-linear variations that may occur in face images later became evident, and explained the fact that those methods tend to fail in non-ideal conditions.

A breakthrough was then made possible by employing features learned through CNN-based networks, e.g. DeepFace \cite{parkhi2015vggdeep} and DeepID \cite{sun2014deepid}. As in previous methods, once the features of two test faces have been computed, a distance measure (typically $\ell_2$ norm) is employed for the verification task: if the distance is below a certain threshold the two test faces are classified as belonging to the same person, otherwise not. 
The loss employed to compute such features is the softmax cross-entropy. Indeed, it was found that the generalization ability could be improved by maximizing inter-class variance and minimizing intra-class variance. Works such as \cite{sun2015deepid3,sun2015deeply} adopted this strategy by accounting for a large margin, in the Euclidean space, between ``contrastive'' embeddings. A further advance was then brought by FaceNet \cite{schroff2015facenet} which introduced the triplet-loss, whereby the distance between the embeddings is evaluated in relative rather than absolute terms. The introduction of the anchor samples in the training process allows to learn embeddings for which the anchor-positive distance is minimized while the anchor-negative distance is maximized. 
Even though this latter work has led to better embeddings, it has been shown that it is oftentimes complex to train \cite{wang2017deep}. 
The focus eventually shifted to the design of new architectures employing metrics other than $\ell_2$ norm to provide more strict margins. In \cite{liu2016large} and \cite{liu2017sphereface} the authors propose to use angular distance metrics to enforce a large margin between negative examples and thus reduce the number of false positives.

In all of the above-mentioned methods, a pre-determined analytical metric is used to compute the distance between two embeddings, and the loss function is designed in order to ensure a large margin (in terms of the employed metric) among the features of negative pairs while compacting the distance among the positive ones. It is important to underline that the chosen metric is a critical aspect in the design of such neural networks. Indeed, a large performance increase has been achieved with the shift from Euclidean to angular distance metrics \cite{deng2019arcface,wang2018cosface}. 

\begin{figure}[tb]
    \centering
    \includegraphics[width=0.6\textwidth]{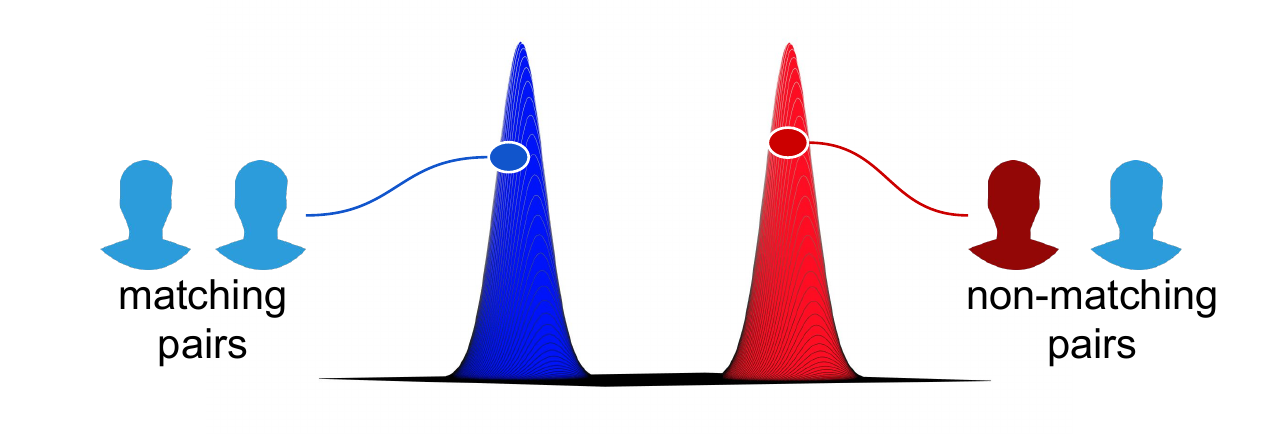}
    \caption[]{The goal of BioMetricNet is to map the input pairs onto target distributions in the latent space. Matching pairs (same user - blue) are mapped to a target distribution whose mean value is far from that of the non-matching pairs (different users - red).}
    \label{fig:goal}
\end{figure}

 In this work we propose a different approach: not only we aim to learn the most discriminative features, but also to jointly learn the best (possibly highly non-linear) metric to compare such features. The only requirement we impose determines how the metric should behave depending on whether the features are coming from matching or non-matching pairs. Specifically, we regularize the metric output such that its values follow two different statistical distributions: one for matching pairs, and the other for non-matching pairs (see Fig. \ref{fig:goal}).
 
The idea of relying on the (empirical) distributions of the feature distances in order to improve their discriminative ability was discussed in \cite{ustinova2016learning}. In the above work, the authors introduce the histogram loss in order to minimize the overlap between the histograms of the distances of matching and non-matching feature pairs, so as to obtain more regularized features. However, while this approach fits well clustering tasks in which one is only interested in relative distances between pairs, it is not suited for the verification problem we are considering in this paper: the decision boundary between the two histograms is highly dependent on the employed dataset and does not generalize across different data distributions.
The approach we follow is rather different: by regularizing a latent space by means of target distributions we \textit{impose} the desired shape (based on a possibly highly non-linear metric) and thus have a known and fixed decision boundary which generalizes across different datasets.
This seminal idea of employing target distributions was first introduced in \cite{RegNet} and \cite{ali2019authnet}. However, it is important to underline that in \cite{RegNet} and \cite{ali2019authnet} it was used to solve a one-against-all classification problem, regularizing a latent space such that the biometric traifts of \textit{a single user} would be mapped onto a distribution, and those of every other possible user onto another distribution, so that a thresholding decision could be used to identify biometric traits belonging to that specific user. The above methods also required a user-specific training of the neural network.

Conversely, besides learning features, the neural network proposed in this paper, which we name BioMetricNet, shapes the decision \textit{metric} such that pairs of similar faces are mapped to a distribution, whereas pair of dissimilar faces are mapped to a different distribution, thereby avoiding user-specific training. This approach has several advantages: i) Since the distributions are known, and generally simple, the decision boundaries are simple, too. This is in contrast with the typical behavior of neural networks, which tend to yield very complex boundaries; ii) If the distributions are taken as Gaussian with the same variance, then a hyperplane is the optimal decision boundary. This leads to a very simple classifier, which learns a complex mapping to a high-dimensional latent space, in a way that mimics kernel-based methods. Moreover, Gaussian distributions are amenable to writing the loss function in closed form; iii) Mapping to known distributions easily enables to obtain confidences for each test sample, as more difficult pairs are mapped to the tails of the distributions. Since in BioMetricNet the distribution of the metric output values is known, the decision threshold can be tuned to achieve the desired level of false alarm rate or genuine acceptance rate. 

The resulting design, employing the best learned metric for the task at hand, allows us to improve over the state-of-the-art also in the case of very challenging datasets, as will be shown in Sec. \ref{sec:performance_comparison}.
We stress that, although in this paper BioMetricNet is applied to faces, the method is general and can be applied to other biometric traits or data types; this is left as future work.

\begin{figure*}[t!]
    \centering
    \includegraphics[width=0.8\linewidth]{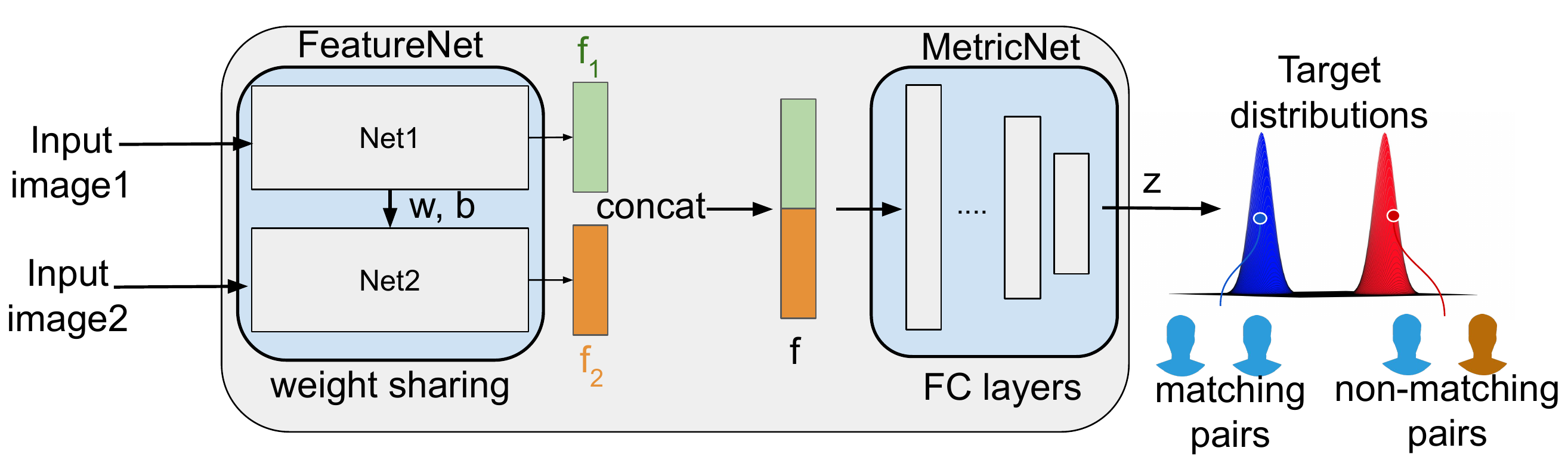}
    \caption[]{BioMetricNet architecture during the training phase. After face detection and alignment, matching and non-matching face pairs are given as an input to the FeatureNet to extract the discriminative face features from the image space $\xv$ into feature vector space $\fv_i \in \mathbb{R}^d$.  The feature vectors are concatenated ${\fv} = [\fv_1 \fv_2] \in \mathbb{R}^{2d}$ and passed to the MetricNet which maps ${\fv}$ onto well-behaved target distributions  $\zv \in \mathbb{R}^p$ in the latent space.}
    \label{fig:proposed_architecture}
\end{figure*}

\section{Proposed Method}

BioMetricNet strives to learn meaningful features of the input faces along with a discriminative metric to be used to compare two sets of facial features. 
More specifically, as depicted in Fig. \ref{fig:proposed_architecture}, BioMetricNet is made of two sub-networks: FeatureNet and MetricNet. The former is a siamese network which processes pairs of input faces ${\xv} = [\xv_1,\xv_2]$ and outputs a pair of facial features ${\fv} = [\fv_1,\fv_2]$ for both matching and non-matching input pairs. 
MetricNet is then employed to map these feature pairs onto a point $\zv$ in a $p$-dimensional space in which a decision is made. These two networks are trained as a single entity to match the desired behavior. Their architecture is described more in detail in Sec. \ref{sec:architecture}.

The novelty of our approach is that we do not impose any predetermined metric between $\fv_1$ and $\fv_2$: the metric is rather learned by MetricNet shaping the decision space according to two target distributions through the loss function, as described in the following.
The loss function forces the value of the learned metric to follow different statistical distributions when applied to matching and non-matching pairs, respectively. 
Although arbitrary target distributions can be employed, a natural choice is to use distributions that have far-enough mass centers, lead to simple decision boundaries, and lend themselves to writing the loss function in a closed form.

For BioMetricNet, let us denote as $\mathbb{P}_m$ and $\mathbb{P}_n$ the desired target distributions for matching and non-matching pairs, respectively. We choose $\mathbb{P}_m$ and $\mathbb{P}_n$ to be multivariate Gaussian distributions over a $p$-dimensional space:
\begin{align}
\mathbb{P}_m = \mathcal{N}(\muv_{m},\Sigmam_{m}),\;\mathbb{P}_n = \mathcal{N}(\muv_{n},\Sigmam_{n}),
\end{align}
where $\Sigmam_{m} = \sigma^2_{m} \mathbb{I}_p$ and $\Sigmam_{n} =  \sigma^2_{n} \mathbb{I}_p$ are  diagonal covariance matrices and $\muv_{m} = \mu_{m} \mathbf{1}^T_p$, $\muv_{n} = \mu_{n} \mathbf{1}^T_p$ are the expected values.
The choice of using Gaussian distributions is a very natural one in this context. Because of the central limit theorem \cite{neal2012bayesian}, the output of fully connected layers tends to be Gaussian distributed.  Moreover, if $\Sigmam_m = \Sigmam_n$, then a linear decision boundary (hyperplane) is optimal for this Gaussian discrimination problem.  Therefore, while in general BioMetricNet can be trained to match arbitrary distributions, in the following we will describe this specific case.
It can also be noted that using different variance for the two distributions would complicate the choice of the parameters, since the optimal variance will be specific to the considered dataset in order to match its intra and inter-class variances.

As said above, in the Gaussian case the loss function can be written in closed form. 
Let us define ${\xv}_m$ and ${\xv}_n$ as the pairs of matching and non matching face images, respectively. In the same way we define ${\fv}_m$ and ${\fv}_n$ as the corresponding features output by FeatureNet. 
MetricNet can be seen as a generic encoding function $H(\cdot)$  of the input feature pairs $\zv = H({\fv})$, where $\zv \in \mathbb{R}^p$, such that $\zv_m \sim \mathbb{P}_m$ if ${\fv} = {\fv}_m$ and $\zv_n \sim \mathbb{P}_n$ if ${\fv} = {\fv}_n$.
As previously described, we want to regularize the metric space where the latent representations $\zv$ lie in order to constrain the metric behavior.
Since the distributions we want to impose are Gaussian, the Kullback-Leibler (KL) divergence between the sample and target distributions can be obtained in closed-form as a function of only first and second order statistics and can be easily minimized. 
More specifically, the KL divergence for multivariate Gaussian distributions can be written as:

\begin{align}
      \mathcal{L}_m &= \frac{1}{2}\left[ \log \frac{|\Sigmam_{m}|}{|\Sigmam_{Sm}|} -p + \mathrm{tr}(\Sigmam_{m}^{-1}\Sigmam_{Sm}) \right. +  \left. (\muv_{m}-\muv_{Sm})^\intercal \Sigmam_{m}^{-1}(\muv_{m}-\muv_{Sm}) \right].
     \label{eq:loss}
\end{align}

where the subscript $S$ indicates the sample statistics. 

Interestingly, since we  only need the first and second order statistics of $\zv$, we can capture this information batch-wise. As will be explained in detail in  Sec. \ref{sec:pairs}, during the training the network is given as input a set of face pairs from which a subset of $b/2$ difficult matching and $b/2$ difficult non-matching face pairs are extracted, being $b$ the batch size. Letting $\Xm \in \mathbb{R}^{b \times r}$ with $r$ the size of a face pair, this results in a collection of latent space points $\Zm \in \mathbb{R}^{b \times p}$ after the encoding. We thus compute first and second order statistics of the encoded representations $\Zm_m,\Zm_n$ related to matching  ($\muv_{Sm}$,$\Sigmam_{Sm}$) and non-matching ($\muv_{Sn}$,$\Sigmam_{Sn}$) input faces respectively.
More in detail, let us denote as $\Sigmam_{Sm}^{(ii)}$ the $i$-th diagonal entry of the sample covariance matrix of $\Zm_m$.
The diagonal covariance assumption  allows us to further simplify \eqref{eq:loss} as:
\begin{align}
    \mathcal{L}_m = \frac{1}{2} \biggr[ \log\frac{\sigma_{m}^{2p}}{\prod_i \Sigmam_{Sm}^{(ii)}} -p& + \frac{\sum_i \Sigmam_{Sm}^{(ii)}}{\sigma^2_{m}} + \frac{||\muv_{m} - \muv_{Sm}||_2}{\sigma^2_{m}} \biggl].
\end{align}

This loss captures the statistics of the matching pairs and enforces the target distribution $\mathbb{P}_m$. For brevity we omit the derivation of $\mathcal{L}_n$ which is obtained similarly.

Then, the overall loss function which will be minimized end-to-end across the whole network (FeatureNet and MetricNet) is given by $\mathcal{L} = \mathcal{L}_m + \mathcal{L}_n$. 

\subsection{Architecture}
\label{sec:architecture}
In the following we  discuss the architecture and implementation strategy of FeatureNet and MetricNet.

\subsubsection{FeatureNet}
The goal of FeatureNet is to extract the most distinctive facial features from the input pairs.
The architectural design of FeatureNet is crucial. In general, one may employ any state-of-the-art neural network architecture able to learn good features. Due to its fast convergence, in our tests we employ a siamese Inception-ResNet-V1 \cite{szegedy2017inception}. The output size of the stem block in Inception-ResNet is $35 \times 35 \times 256$, followed by 5 blocks of Inception-ResNet-A, 10 blocks of Inception-ResNet-B and 5 blocks of Inception-ResNet-C. At the bottom of the network we employ a fully connected layer with output size equal to the feature vector dimensionality $d$.  The employed dropout rate is $0.8$. 
The pairs of feature vectors $\fv_1$ and $\fv_2$ in output of FeatureNet are concatenated resulting in $\fv = [\fv_1 \fv_2] \in \mathbb{R}^{2d}$ and given as input to MetricNet.
\subsubsection{MetricNet}
The goal of MetricNet is to learn the best metric based on feature vector $\fv$ and to map it onto the target distributions in the latent space.
MetricNet consists of $7$ fully connected layers with ReLU activation functions at the output of each layer. At the last layer, no activation function is employed. The input size of MetricNet is $2d$, the first fully connected layer has an output size equal to $2d$, the output size keeps decreasing gradually by a factor of 2 with the final layer having an output size equal to the latent space dimensionality $p$. 

We also highlight that MetricNet, by taking as input ${\fv} = [\fv_1,\fv_2]$, allows us to model any arbitrary nonlinear correlations between the feature vectors. Indeed, the use of an arbitrary combination of the input features entries has been proven to be highly effective, see e.g. \cite{chen2017once,held2016learning}.



\subsection{Pairs Selection during Training}
\label{sec:pairs}
For improved convergence, BioMetricNet selects the most difficult matching and non-matching pairs during training, i.e., those far from the mean values of the target distributions and close to the threshold. For each mini-batch, at the end of the forward pass we select the subset of matching pairs whose output $\zv_m$ is sufficiently distant from the mass center of $\mathbb{P}_m$, i.e. $|| \zv_m- \muv_{m}||_{\infty}\geq 2\sigma_m$.
Similarly, for the non-matching pairs we select those which result in a $\zv_n$ such that $|| \zv_n- \muv_{n}||_{\infty}\geq 2\sigma_n$.
Then, in the backward pass we minimize the loss over a subset of $b/2$ difficult matching and $b/2$ difficult non-matching pairs with $b$ being the mini-batch size of the selected difficult pairs. In order to have a stable training, the backward pass is executed only when we are able to collect $b/2$ difficult matching and $b/2$ difficult non-matching pairs, else the mini-batch is discarded. 

The rationale behind this choice comes from the result of the latent space regularization. Indeed, as one traverses the latent space from $\muv_m$ towards $\muv_n$, one moves from very similar face pairs to very dissimilar ones. Points close to the threshold can be thought of as representing pairs for which the matching/non-matching uncertainty is high. 
As the training proceeds, at every new epoch the network improves the mapping of the ``difficult'' pairs as it is trained on pairs for which it is more difficult to determine whether they represent a match or not. 

\begin{figure}[t]
    \centering
    \includegraphics[width=0.6\columnwidth]{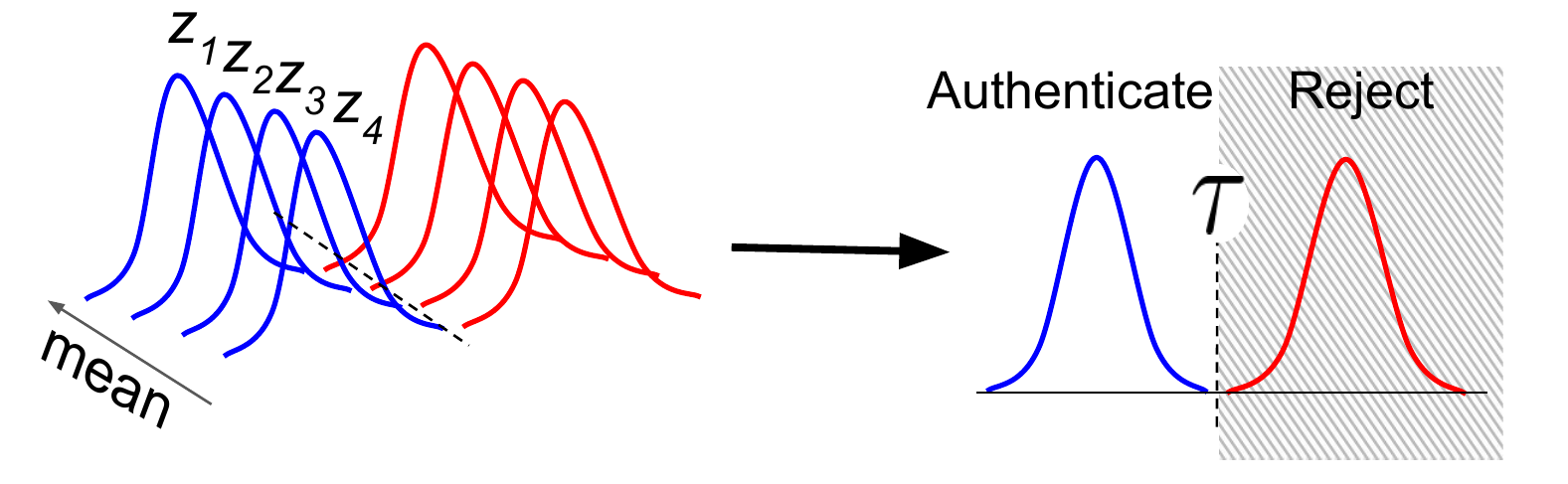}
    
    \caption[]{During the testing phase, we obtain the latent vectors of the input image pair and its three horizontal flips. For all experiments, the final latent space vector is calculated as $ \bar{\zv} = \frac{1}{4}\left (\sum_{i=1}^4{\zv_i}\right ) $. Pairs are classified as matching and non-matching by comparing $\bar{\zv}$ with a threshold $\tau$.
    }
    \label{fig:BioMetricNet-M}
\end{figure}

\subsection{Authentication}
In the testing phase, a pair of images are passed through the whole network in order to compute the related metric value $\zv$. Then, a decision is made according to this value. As said, for our choice of target distributions a hyperplane can be used for the optimal decision, i.e., we can use the test 
\begin{equation}
    \label{eq:test}
(\muv_m - \muv_n)^T\zv \lessgtr (\muv_m - \muv_n)^T(\muv_m + \muv_n)/2.
\end{equation}
For $p=1$, this boils down to comparing the scalar $\zv$ with a threshold $\tau=(\mu_m+\mu_n)/2$.

\begin{figure}[t!]
    \centering
    \includegraphics[width=0.8\columnwidth]{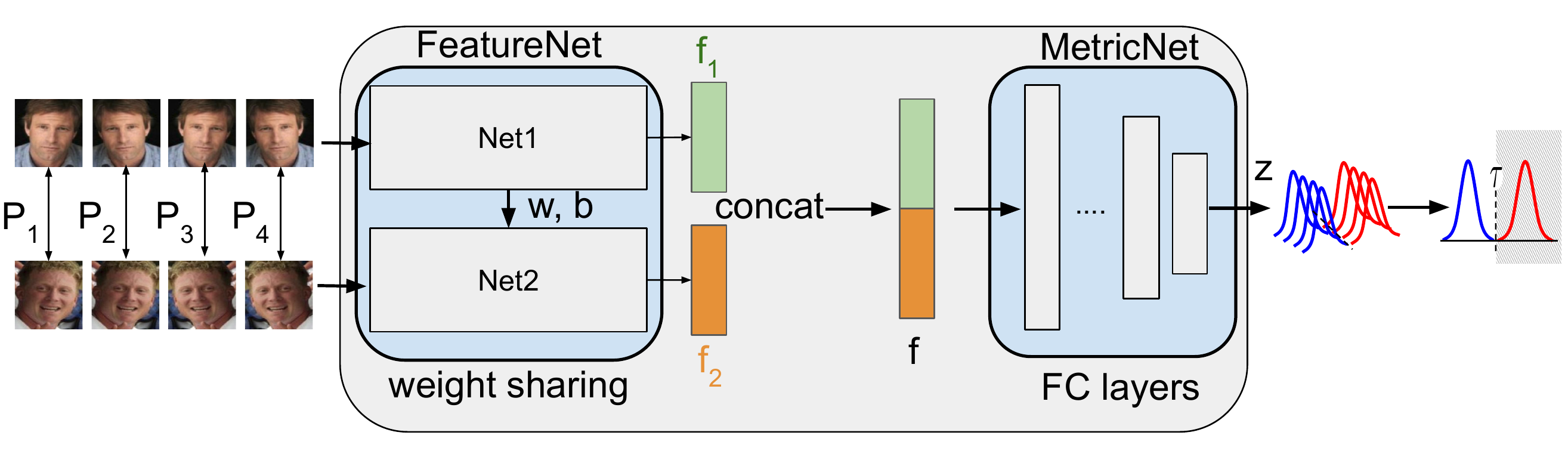}
    \caption[]{BioMetricNet architecture during the testing phase. Given a pair of images to be tested, after face detection and alignment, by accounting for all the possible horizontal flip combinations, we obtain $4$ image pairs, i.e. ${P_1,P_2,P_3}$ and $P_4$. The latent vectors of the corresponding pairs are computed and aggregated to $\bar{\zv}$ and compared with a threshold $\tau$.}
    \label{fig:proposed_testing}
\end{figure}

However, we consider an improved approach which is able to capture additional information: we use flipped images to compute supplementary features as done in the recent literature \cite{liu2017sphereface,wang2018cosface}. 
Namely, given an image pair, we compute the metric output $\zv$ for the original image pair as well as the $3$ pairs resulting from the possible combinations of horizontally flipped and non-flipped images. We employ a horizontal flip defined as $(x, y)\xrightarrow{} (width - x - 1, y)$. We thus obtain 4 metric values.
Then, the decision is performed on a value $ \bar{\zv} = \frac{1}{4}\left(\sum_{i=1}^4{\zv_i}\right)$, where $\zv_i$ is the metric output corresponding
to the $i$-th image flip combination, see Fig. \ref{fig:BioMetricNet-M}. 
The expected value of $\bar{\zv}$ in case of matching and non-matching pairs is still equal to $\muv_m$ and $\muv_n$, respectively.
Therefore, the test (\ref{eq:test}) will still be valid on $\bar{\zv}$. Fig. \ref{fig:proposed_testing} depicts BioMetricNet during the authentication phase. P1 represents the input image pair, and P2, P3, and P4 represent the three horizontal flips.

\section{Experiments}

\subsection{Experimental Settings}
The network is trained with Adam optimizer \cite{kingma2014adam} using stochastic gradient descent \cite{lecun1989backpropagation,rumelhart1988learning}. Each epoch consists of $720$ people with each person having a minimum of $5$ images to ensure enough matching and non-matching pairs. We set the batch size $b$ to 220 difficult pairs to obtain statistically significant first and second order statistics. Each batch is balanced, i.e. it contains half matching and half non-matching pairs.
The initial learning rate is set to $0.01$ with an exponential decay factor of $0.98$ after every $5$ epochs. In total, the network is trained for $500000$ iterations. Weight decay is set to $2 \times 10^{-4}$. We further employ dropout with a keep probability value equal to $0.8$. All experiments are implemented in TensorFlow \cite{abadi2016tensorflow}. 
For the augmentation horizontal flips of the images are taken.

\subsection{Preprocessing}
For preprocessing we follow the strategies adopted by most recent papers in the field \cite{deng2019arcface,wang2018cosface,liu2017sphereface}. For both the training and testing datasets, we employ MTCNN \cite{zhang2016joint} to generate normalized facial crops of size $160 \times 160$ with face alignement based on five facial points. As a final step, the images are mean normalized and constrained in the range $[-1,1]$ as done in  \cite{wang2018cosface,liu2017sphereface,deng2019arcface}.

\subsection{Datasets}
\subsubsection{Training} The training datasets are those commonly used in recent works in the field. More in detail, we use different datasets for training and testing phases. The datasets we employ during training are Casia \cite{yi2014casia} (0.49M images having 10k identities)
and MS1M-DeepGlint (3.9M images having 87k identities)  \cite{ms1mv2}. 


\begin{figure}[tb]
    \centering
    \includegraphics[width=0.5\columnwidth]{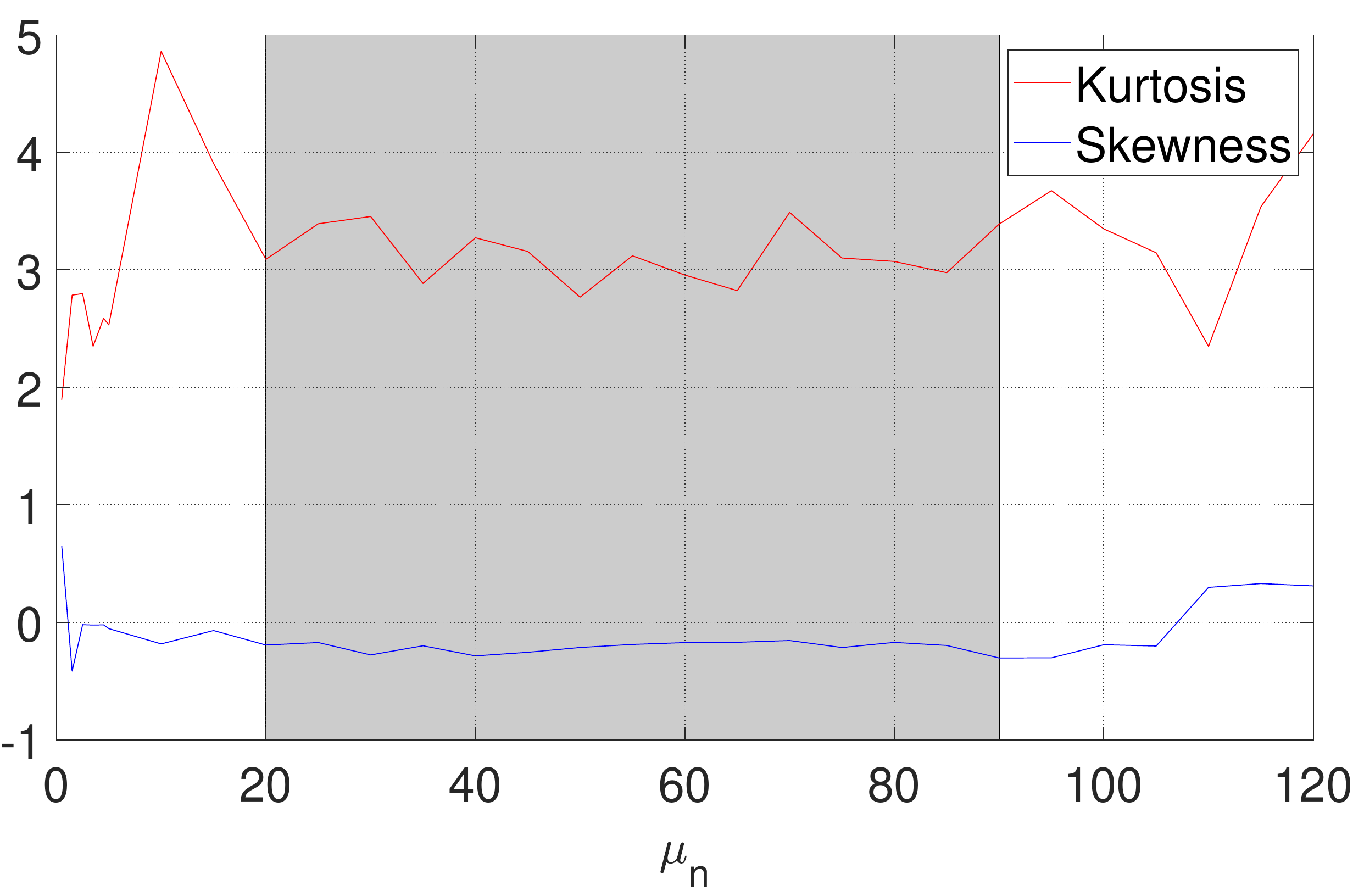}
    
    \caption[]{Kurtosis and skewness of the latent space metric on LFW when $\mathbb{P}_n = \mathcal{N}(w,1)$ where $w=[0.5,120]$, and $\mathbb{P}_m = \mathcal{N}(0,1)$. If the means of the two distribution are too far apart the training process becomes unstable, hence it affects the kurtosis and skewness of the imposed distributions.}
    \label{fig:acc_comp}
\end{figure}

\subsubsection{Testing} 
BioMetricNet, in the current setting, has been developed for 1:1 verification in a face authentication scenario, particularly, when there is a single image template per subject. Therefore, BioMetricNet has been validated on 6 popular unconstrained face datasets for 1:1 verification, excluding large scale datasets like MegaFace \cite{kemelmacher2016megaface} and IJB \cite{maze2018iarpa} used for set-based face recognition, i.e., deciding whether two {\em  sets} of images of a face belong to the same person or not.

Labeled Faces in the Wild (LFW) \cite{huang2008lfw} and YouTube Faces (YTF) \cite{wolf2011ytf} are the most commonly used datasets for unconstrained face verification on images and videos. LFW consists of 13233 face images collected from 5749 people. YTF consists of 3425 videos of 1595 people. Latest deep learning models for face verification are powerful enough to
achieve almost perfect accuracy on LFW and YTF, making the related results not very informative.
For detailed insights we further test BioMetricNet on more challenging datasets such as Cross-Age LFW (CALFW) \cite{zheng2017calfw} which is constructed by selecting 3000 positive face pairs with age gap from LFW to add the aging process to intraclass variance, and Cross-Pose LFW (CPLFW) \cite{zheng2018cplfw} which is constructed from 3000 face pairs of LFW with pose difference to add pose variation to intra-class variance. Finally, we evaluate our method on Celebrity dataset in frontal and profile views (CFP) \cite{sengupta2016cfp} having 500 identities with 7000 images, and in-the-wild age database (AgeDB) \cite{moschoglou2017agedb} containing 16488 images of 568 identities. For all the datasets, we report the results for 6000 pairs of testing images and videos having 3000 matching and 3000 non-matching pairs. For reporting the performance we follow the standard protocol of \textit{unrestricted with labeled outside data} as done in \cite{schroff2015facenet,deng2019arcface,wang2018cosface}.

\subsection{Effect of Feature Vector Dimensionality} We explored the effect of different dimensionality of the feature vector by fixing $p=1$, and varying $d$, see Tab. \ref{table:FeatureVectorDimensionality}. It can be observed that small values of $d$ are not sufficient to capture the most  discriminative facial features. On the other hand, a too large feature space ($1024$) causes overfitting and thus a performance drop.
We picked the best value, i.e. $d=512$, since in our experiments this choice leads to the highest accuracy.

\subsection{Effect of Latent Space Dimensionality}
In order to select the optimal latent space size we explored different dimensionalities by fixing $d=512$. The results are shown in Tab. \ref{table:FeatureVectorDimensionality}. In this case, as a general behavior it can be observed that an increase in $p$ leads to a performance drop. 


Since $p$ affects the number of parameters at the very bottom of MetricNet (an FC network), its choice strongly affects the overall performance. We conjecture that large values of $p$ might be beneficial for very complex dataset, for which the amount of training data is typically large. Indeed, samples in a higher dimensional latent space are generally more linearly separable. This is even more important when the number of data points is very large.
On the other hand, too large values of $p$ might lead to a performance drop as it becomes difficult to learn a mapping onto large latent space.
From Tab. \ref{table:FeatureVectorDimensionality} it can be seen that for most of the datasets, $p=1$ is sufficient.
On the other hand, for CFP-FP it can be seen seen that the highest accuracy (even though by a small amount) is reached for $p=16$. In this case, a higher latent space dimensionality provides room to achieve a better separation. Since $p=1$ provides optimal or close to optimal results in all cases, we choose this value for the experiments.

\begin{table}[tb]
 \centering
 \caption[]{Accuracy (\%) for different feature vector $d$ and latent vector $p$ dimensionality. Highest accuracy is obtained for the feature vector of size $d = 512$ and for $p=1$ }
\resizebox{1.0\columnwidth}{!}{
\begin{tabular}{ C{2cm}| C{1.4cm}|C{1.4cm}| C{1.4cm} | C{1.4cm} ? C{1.4cm}|C{1.4cm}| C{1.4cm} | C{1.4cm}  }
\hline
\textbf{Dataset}   & $d  = 128$     & $d  = 256$  & $d  = 512$ & $d  = 1024$ & $p = 1$   & $p  = 3$     & $p = 8$  & $p  = 16$ \\ \Xhline{3\arrayrulewidth}

LFW  & 99.47 &  99.51 & \textbf{99.80} & 99.63 & \textbf{99.80} & 99.75 &  99.74 & 99.72 \\ 

YTF  & 97.57 &  97.76 & \textbf{98.06} & 98.0  & \textbf{98.06} & 97.85 &  97.73 & 97.76 \\ 

CALFW  & 96.48 &  96.59 & \textbf{97.07} & 96.78 & \textbf{97.07} & 97.02 & 96.92 & 96.93 \\ 

CPLFW  & 94.89 &  94.81 & \textbf{95.60} & 95.25 & \textbf{95.60} & 95.57 &  95.13 & 95.43 \\ 

CFP-FP  & 99.01 &  99.08 & \textbf{99.35} & 99.25 & 99.35 & 99.33 &  99.33 & \textbf{99.47} \\ 

 \Xhline{3\arrayrulewidth}
  \end{tabular}}

\label{table:FeatureVectorDimensionality}
\end{table}

\subsection{Parameters of Target Distributions}

\begin{table*}[t]
 \centering
 \caption[]{\textbf{Verification accuracy \%} of different methods on LFW, YTF, CALFW, CPLFW, CFP-FP and AgeDB. BioMetricNet achieves state-of-the-art results for YTF, CALFW, CPLFW, CFP-FP, and AgeDB and obtains similar accuracy to the state-of-the-art for LFW}
\resizebox{\textwidth}{!}{
\begin{tabular}{ C{4cm}| C{2cm}|C{2cm}| C{2cm}| C{2cm}| C{2cm}| C{2cm}| C{2cm} }
\hline
 \textbf{Method}   &  $\mathbb{\#}$ \textbf{Image}   & \textbf{LFW}     & \textbf{YTF} & \textbf{CALFW} & \textbf{CPLFW} & \textbf{CFP-FP} & AgeDB  \\ \Xhline{3\arrayrulewidth}

SphereFace \cite{liu2017sphereface} & 0.5M & 99.42 &  95.0& 90.30 & 81.40 & 94.38 & 91.70 \\ 

SphereFace+ \cite{liu2018learning} & 0.5M & 99.47 &  - & - & - & - & - \\ 

FaceNet \cite{schroff2015facenet} & 200M & 99.63 &  95.10 & - & - & - & 89.98 \\

VGGFace \cite{parkhi2015vggdeep} & 2.6M & 98.95 &  97.30& 90.57 & 84.00 & - & - \\

DeepID \cite{sun2014deepid} & 0.2M & 99.47 &  93.20 & - & - & - & - \\

ArcFace \cite{deng2019arcface} & 5.8M & \textbf{99.82} &  98.02 & 95.45 & 92.08 & 98.37 & 95.15\\ 

CenterLoss \cite{wen2016centerloss} & 0.7M & 99.28 &  94.9& 85.48 & 77.48 & - & - \\

DeepFace \cite{taigman2014deepface} & 4.4M & 97.35 &  91.4& - & - & - & - \\

Baidu \cite{liu2015baidu} & 1.3M & 99.13 &  - & - & - & - & - \\

RangeLoss \cite{zhang2017rangeloss} & 5M & 99.52 &  93.7& - & - & - & - \\

MarginalLoss \cite{deng2017marginalloss} & 3.8M & 99.48 &  95.98& - & - & - & - \\ 

CosFace \cite{wang2018cosface} & 5M & 99.73 &  97.6& - & - & 95.44 & - \\ 


BioMetricNet & 3.8M & \textbf{99.80} &  \textbf{98.06}& \textbf{97.07} & \textbf{95.60} & \textbf{99.35} & \textbf{96.12} \\ 
\Xhline{3\arrayrulewidth}
\end{tabular}}
\label{table:verification_accuracy}
\end{table*}


In this section, we perform an experiment to explore the behavior of different parameters of the target distributions. At first, let us recall that we set the two distributions to have the same variance $\sigma_m=\sigma_n=\sigma$. This allows us to have only a single free parameter, i.e. the ratio $(\mu_m - \mu_n)/\sigma$,  
affecting our design, in terms of how far apart we place the distributions compared to the chosen variance.  Without loss of generality, this can be tested by setting the distributions to be $\mathbb{P}_m = \mathcal{N}(0,1)$ and $\mathbb{P}_n = \mathcal{N}(w,1)$, where $w=[0.5,120]$. 
From now on if not differently specified we will consider $p=1$ and $d=512$.

In more detail, in Fig. \ref{fig:acc_comp} we show the skewness and kurtosis of the latent representation as a function of $w$ for LFW dataset. It can be seen that in the region corresponding to $20 \le w \le 90$ the skewness and kurtosis are close to 0 and 3, respectively, and the accuracy is high, showing that the training indeed converges to Gaussian distributions.  We eventually choose $\mu_m=0$ and $\mu_n=40$ to keep the distributions sufficiently far apart from each other. Further, if the difference between $\mu_m$ and $\mu_n$ is too large (e.g. $\mu_  
n > 90$), the training process becomes unstable and the distributions become far from Gaussian.

\subsection{Performance Comparison}
\label{sec:performance_comparison}

Tab. \ref{table:verification_accuracy} reports the maximum verification accuracy obtained for different methods on several datasets. For YTF and LFW as reported in Tab. \ref{table:verification_accuracy}, it can be observed that BioMetricNet achieves higher accuracy with respect to other methods. In particular, it achieves an accuracy of 98.06\% and  99.80\% for YTF and LFW datasets respectively. On these two datasets, ArcFace obtains a comparable accuracy. 

For a more in-depth comparison we further test BioMetricNet on more challenging datasets, i.e. CALFW, CPLFW, CFP-FP and AgeDB. State-of-the-art results on these datasets are far from the ``almost perfect" accuracy we previously observed. 
In Tab. \ref{table:verification_accuracy} we compare the verification performance for these datasets. As can be observed, BioMetricNet significantly outperforms the baseline methods (CosFace, ArcFace, and SphereFace). For CPLFW, BioMetricNet achieves an accuracy of 95.60\% obtaining an error rate that is 3.52\% lower than previous state-of-the-art results, outperforming ArcFace by a significant margin. 
For CALFW, BioMetricNet achieves an accuracy of 97.07\% which is 1.62\% lower than previous state-of-the-art results. For CFP dataset BioMetricNet achieves an accuracy of 99.35\% lowering the error rate by about 1\% with respect to ArcFace. Finally, for AgeDB BioMetricNet achieves an accuracy of 96.12\% lowering the error rate by about 1\% compared to ArcFace.

To summarise, when compared to state-of-art approaches BioMetricNet consistently achieves higher accuracy, proving that, by learning the metric to be used to compare facial features in a regularized space, the discrimination ability of the network is increased. This becomes more evident on more challenging datasets where the gap from perfect accuracy is larger.

\begin{figure}[t]
\centering
\includegraphics[width=0.5\columnwidth]{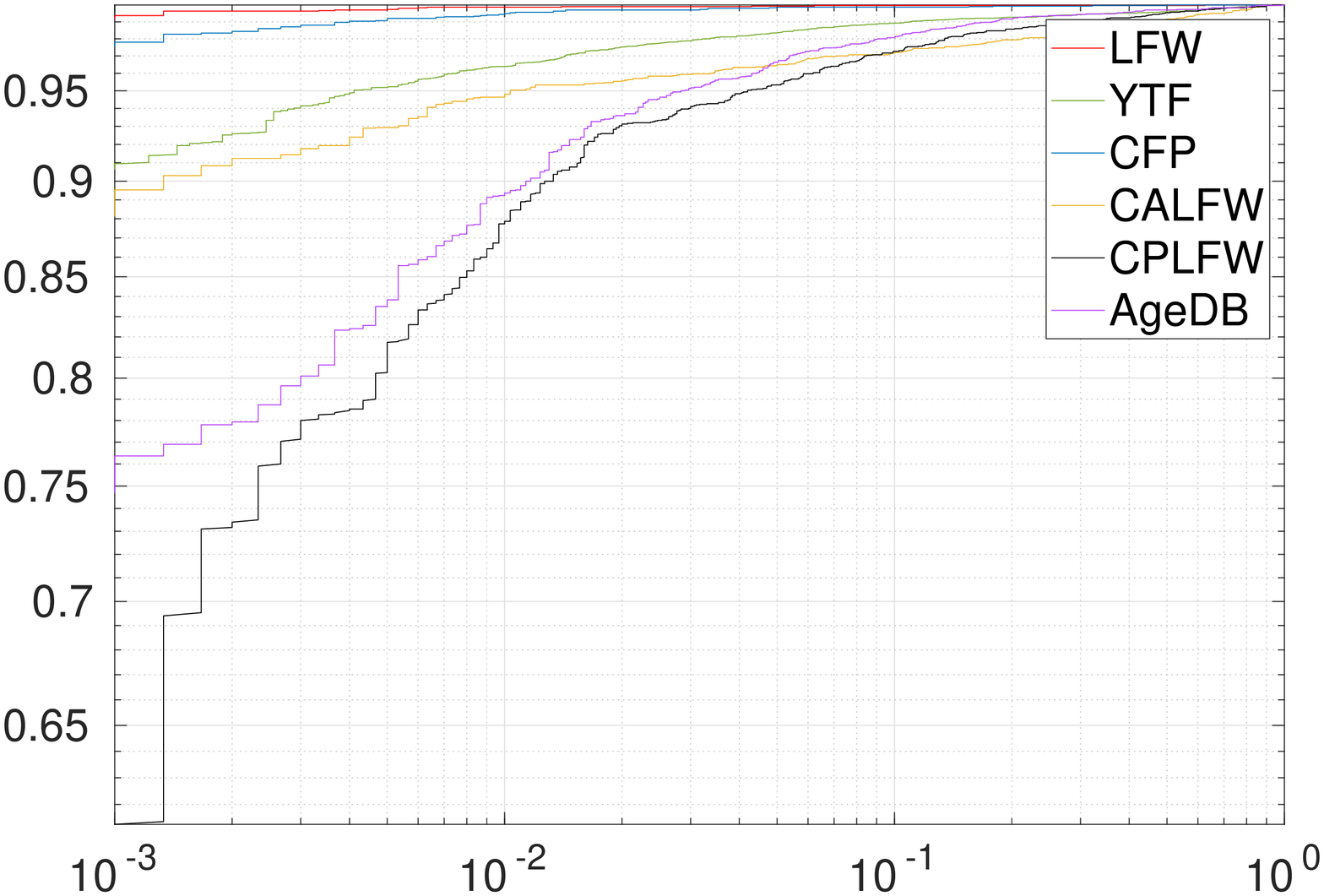}
\caption[]{ROC curve of BioMetricNet on LFW, YTF, CFP, CALFW, CPLFW and AgeDB.}
\label{fig:ROC}
\end{figure}

\begin{table}[t]
 \centering
 \caption[]{GAR obtained for LFW, YTF, CFP, CALFW, CPLFW and AgeDB at FAR=\{${10^{-2}},{10^{-3}}$\}}
\resizebox{0.6\columnwidth}{!}{
\begin{tabular}{ C{2cm}| C{3.2cm}|C{3.2cm}}
\hline
\textbf{Dataset}    & GAR@$\mathbf{10^{-2}}$FAR\%   & GAR@$\mathbf{10^{-3}}$FAR\%    \\ \Xhline{3\arrayrulewidth}

LFW & 99.87 & 99.20  \\ 
 
YTF & 96.93 & 90.87   \\ 

CALFW & 94.63 & 88.13   \\ 

CPLFW & 87.73 & 61.27  \\ 

CFP-FP & 99.43 & 97.57  \\ 

AgeDB-30 & 89.23 & 74.70  \\ 

 \Xhline{3\arrayrulewidth}
  \end{tabular}}
\label{table:ROC}
\end{table}

\subsection{ROC Analysis}
\label{sec:ROC_analysis}
The Receiver Operating Characteristic (ROC) analysis of BioMetricNet is illustrated in Fig. \ref{fig:ROC}. This curve depicts the Genuine Acceptance Rate (GAR), namely the relative number of correctly accepted matching pairs as function of the False Acceptance Rate (FAR), the relative number of incorrectly accepted non-matching pairs. Furthermore, we report the GAR at different FAR values, namely FAR=\{${10^{-2}},{10^{-3}}$\} in Tab. \ref{table:ROC}. By means of the ROC, we can analyze how the verification task solved by BioMetricNet generalizes across different datasets. 
 It is immediate to notice that, as a result of clear separation and low contamination of the area between the matching and non-matching distributions, high GARs are obtained at low FARs. This is generally true at different ``complexity'' levels as exposed by the considered datasets.
 More in detail, for LFW at FAR=${10^{-2}}$ and FAR=$10^{-3}$ high GARs of 99.87\% and 99.20\%  are obtained, see Tab. \ref{table:ROC}. For YTF at FAR=${10^{-2}}$ and FAR=$10^{-3}$,  GARs of 96.93\% and 90.87\% are obtained. The same behavior can be observed for CFP. For challenging datasets of CALFW CPLFW and AgeDB  it can be observed that the ROC curves obtained are comparatively lower compared to LFW, YTF, and CFP.  The GARs at FAR=${10^{-2}}$ and FAR=$10^{-3}$ comes to be 94.63\% and 88.13\% for CALFW, 87.73\% and 61.27\% for CPLFW and 89.23\% and 74.70\% for AgeDB respectively. 

\begin{figure*}[t]
\centering
\begin{subfigure}{.19\textwidth}
  \centering
  \includegraphics[width=\linewidth]{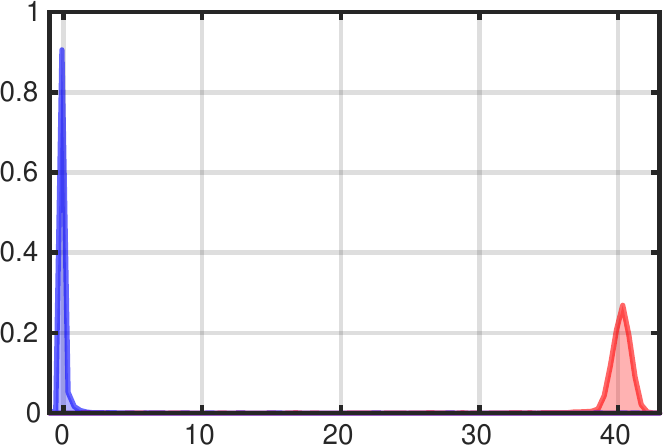}
  \caption[]{{LFW}}
  \label{fig:sub1-1}
\end{subfigure}
\begin{subfigure}{.19\textwidth}
  \centering
  \includegraphics[width=\linewidth]{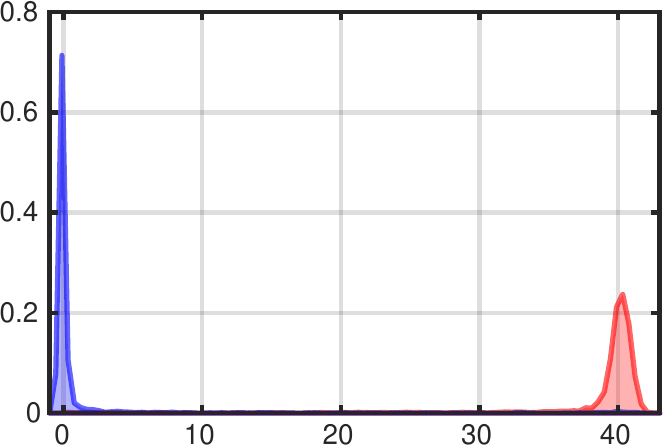}
  \caption[]{YTF}
  \label{fig:sub2-1}
\end{subfigure}
\begin{subfigure}{.19\textwidth}
  \centering
  \includegraphics[width=\linewidth]{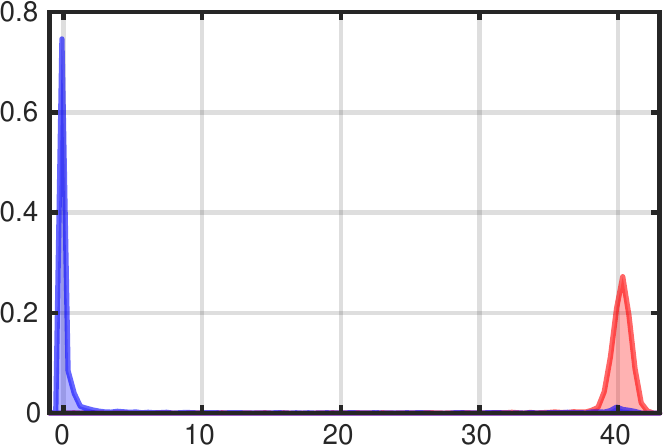}
  \caption[]{CALFW}
  \label{fig:sub4-1}
\end{subfigure}
\begin{subfigure}{.19\textwidth}
  \centering
  \includegraphics[width=\linewidth]{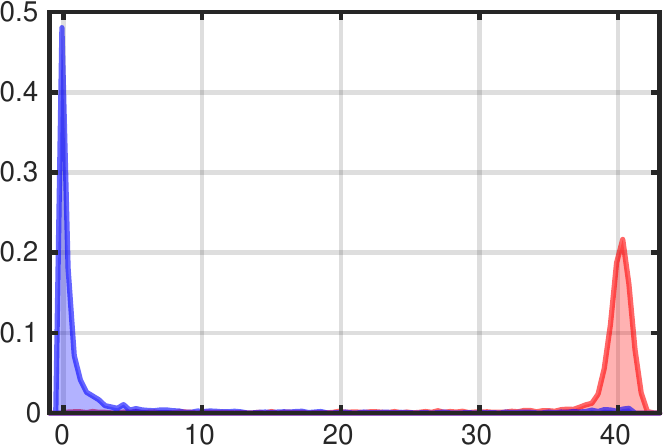}
  \caption[]{CPLFW}
  \label{fig:sub5-1}
\end{subfigure}
\begin{subfigure}{.19\textwidth}
  \centering
  \includegraphics[width=\linewidth]{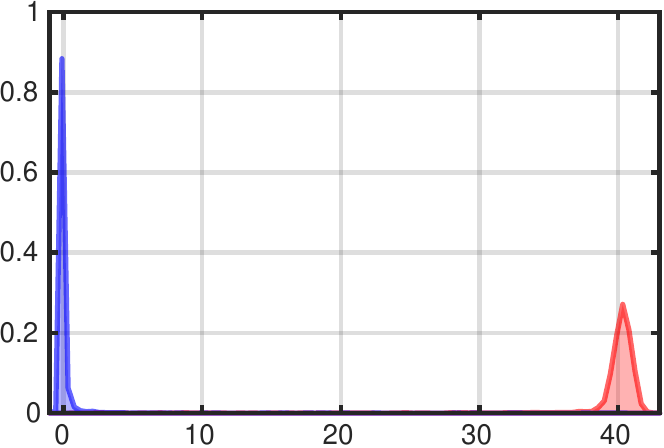}
  \caption[]{CFP-FP}
  \label{fig:sub3-1}
\end{subfigure}
\caption[]{Histogram of $\zv$ decision statistics of BioMetricNet matching and non-matching pairs from (a) LFW; (b) YTF; (c) CALFW; (d) CPLFW; (e) CFP-FP. Blue area indicates matching pairs while red indicates non-matching pairs.}
\label{fig:hist_distibutions-1}
\end{figure*}

\begin{figure*}[t]
\centering
\begin{subfigure}{.19\textwidth}
  \centering
  \includegraphics[width=\linewidth]{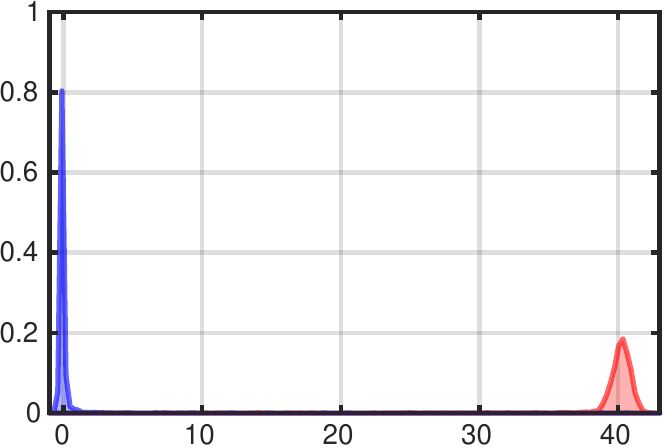}
  \caption[]{{LFW}}
  \label{fig:sub1}
\end{subfigure}
\begin{subfigure}{.19\textwidth}
  \centering
  \includegraphics[width=\linewidth]{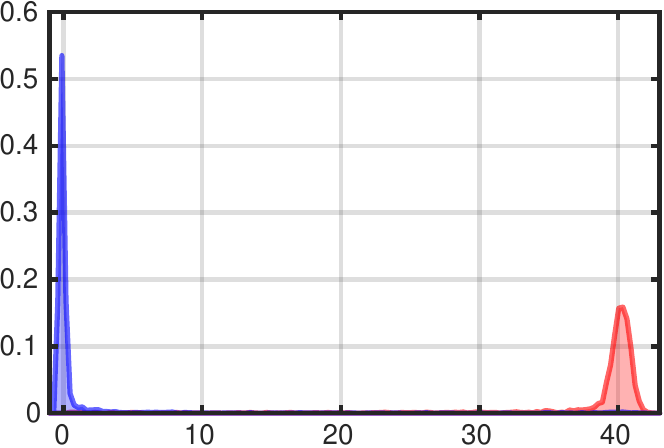}
  \caption[]{YTF}
  \label{fig:sub2}
\end{subfigure}
\begin{subfigure}{.19\textwidth}
  \centering
  \includegraphics[width=\linewidth]{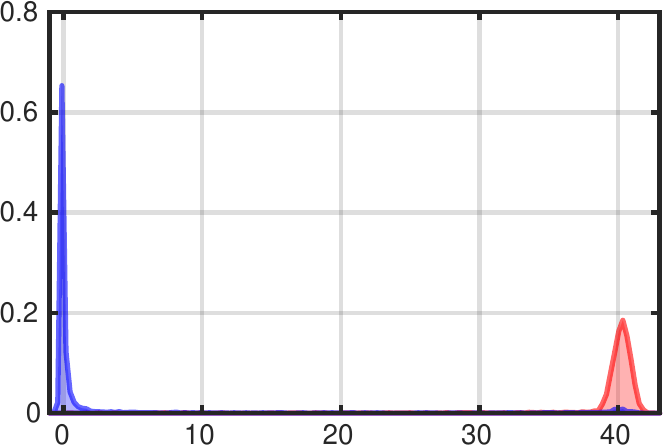}
  \caption[]{CALFW}
  \label{fig:sub4}
\end{subfigure}
\begin{subfigure}{.19\textwidth}
  \centering
  \includegraphics[width=\linewidth]{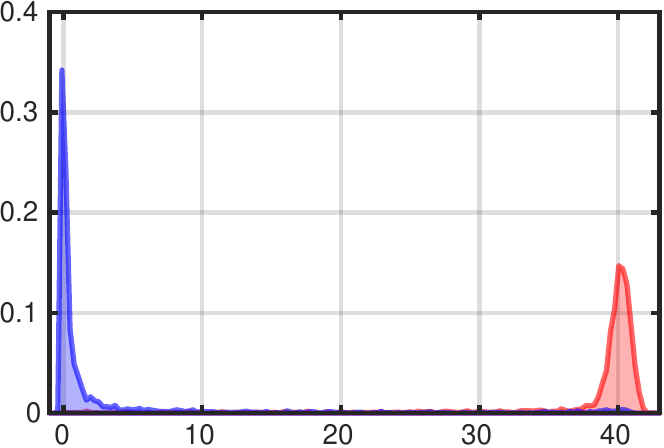}
  \caption[]{CPLFW}
  \label{fig:sub5}
\end{subfigure}
\begin{subfigure}{.19\textwidth}
  \centering
  \includegraphics[width=\linewidth]{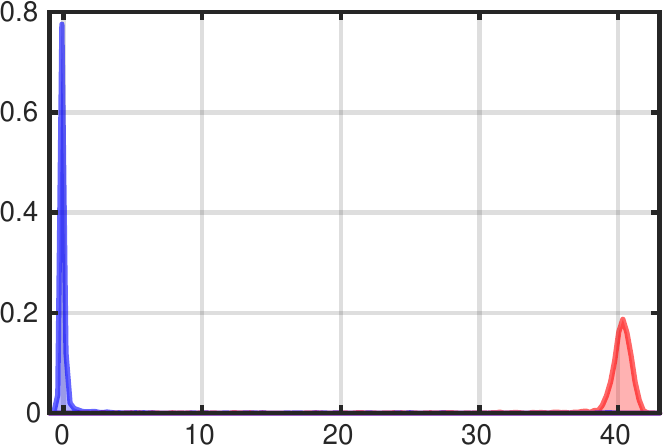}
  \caption[]{CFP-FP}
  \label{fig:sub3}
\end{subfigure}
\caption[]{Histogram of $\bar{\zv}$ decision statistics of BioMetricNet matching and non-matching pairs from (a) LFW; (b) YTF; (c) CALFW; (d) CPLFW; (e) CFP-FP. Blue area indicates matching pairs while red indicates non-matching pairs.}
\label{fig:hist_distibutions}
\end{figure*}

\subsection{Analysis of Metrics Distribution}
BioMetricNet closely maps the metrics for matching and non-matching pairs onto the imposed target Gaussian distributions. To analyze more in depth the effects of the latent space regularization, we depict the histograms of $\zv$ and $\bar{\zv}$ computed over different test datasets, in Fig. \ref{fig:hist_distibutions-1} and Fig. \ref{fig:hist_distibutions} respectively.
At first, it can be noticed that for both $\zv$ and $\bar{\zv}$ the proposed regularisation is able to shape the latent space as intended by providing Gaussian-shaped distributions. Observing the histograms of $\zv$ and $\bar{\zv}$, it can be noticed that for all the datasets BioMetricNet very effectively separates matching and non-matching pairs.

Concerning non-matching pairs, the distributions of $\zv$ are indeed Gaussian with the chosen parameters. For matching pairs, it can be observed that the $\zv$ score has the correct mean, but tends to have a lower variance than the target distribution. A possible explanation is that matching and non-matching pairs exhibit different variability, so it is difficult to match them to distributions with the same variance. Indeed, for a fixed number of persons, the number of possible non-matching pairs is much larger than the number of possible matching pairs. Moreover, the KL divergence is not symmetric and the chosen loss tends to promote sample distributions with a smaller variance than the target one, rather than with a larger variance. Hence, a solution where matching pairs have a smaller variance than the target distribution is preferred with respect to a solution where non-matching pairs have a larger variance than the target distribution. We can also observe that for more difficult datasets, like CALFW and CPLFW, the distribution obtained for matching pairs has heavier tails than the target distribution.

The histogram for $\bar{\zv}$ scores in Fig. \ref{fig:hist_distibutions} shows that the variance of both matching and non-matching pairs is slightly reduced with respect to that of $\zv$. Since reduced variance means increased verification accuracy, this justifies using $\bar{\zv}$ over $\zv$.
Furthermore, the decision boundary we are using depends only on mean values, which are preserved, and thus it is not affected by the slight decrease in variance.

\section{Conclusions}
We have presented a novel and innovative approach for unconstrained face verification mapping learned discriminative facial features onto a regularized metric space, in which matching and non-matching pairs follow specific and well-behaved distributions. The proposed solution, which does not impose a specific metric, but allows the network to learn the best metric given the target distributions, leads to improved accuracy compared to the state of the art. In BioMetricNet distances between input pairs behave more regularly, and instead of learning a complex partition of the input space, we learn a complex metric over it which further enables the use of much simpler boundaries in the decision phase.
With extensive experiments, on multiple datasets with several state-of-the-art benchmark methods, we showed that BioMetricNet consistently outperforms other existing techniques.
Future work will consider BioMetricNet in the context of 3D face verification and adversarial attacks. Moreover, considering the slight mismatch between metric distributions and target distributions, it is worth investigating if alternative parameter choices for the target distributions can lead to improved results.

\section{Acknowledgment}
This work results from the research cooperation with Sony R\&D Center Europe Stuttgart Laboratory 1.
\newpage

%

\bibliographystyle{splncs}
\bibliography{main}

\end{document}